\documentclass{article} 
\usepackage{jmlr2e}
\usepackage{hyperref}
\usepackage{url}
\usepackage{amsmath}
\usepackage{graphicx}
\usepackage{subcaption}
\usepackage{siunitx}
\usepackage{booktabs}
\usepackage{float}
\usepackage{multirow}

\begin{document}
\title{Fast ES-RNN: A GPU Implementation of the ES-RNN Algorithm}
\author{\name Andrew Redd\thanks{All three authors contributed equally}\\
        \email aredd@andrew.cmu.edu\\
       \addr Carnegie Mellon University\\
       Pittsburgh, PA, USA
       \AND
       \name Kaung Khin\footnotemark[1]\\
        \email kkhin@andrew.cmu.edu\\
       \addr Carnegie Mellon University\\
       Pittsburgh, PA, USA
       \AND
       \name Aldo Marini\footnotemark[1]\\
        \email amirinim@andrew.cmu.edu\\
       \addr Carnegie Mellon University\\
       Pittsburgh, PA, USA}

\maketitle
\begin{abstract}%
Due to their prevalence, time series forecasting is crucial in multiple domains. We seek to make state-of-the-art forecasting fast, accessible, and generalizable. ES-RNN is a hybrid between classical state space forecasting models and modern RNNs that achieved a 9.4\% sMAPE improvement in the M4 competition. Crucially, ES-RNN implementation requires per-time series parameters. By vectorizing the original implementation and porting the algorithm to a GPU, we achieve up to 322x training speedup depending on batch size with similar results as those reported in the original submission. Our code can be found at: \textit{https://github.com/damitkwr/ESRNN-GPU}
\end{abstract}
\section{Introduction}
\label{sec:intro}

Time series are ubiquitous. Data from diverse sources such as sensors, financial markets, demographics, audio and Uber rides have a time component. Time series' prevalence implies a need to develop powerful forecasting methods. However, there are two major reasons why popular machine learning algorithms have failed to outperform as they have in other domains. First, the time component breaks the independent and identically distributed assumption by introducing auto-correlation. Machine learning has struggled overcome this assumption gap and offer step-wise improvement over pure statistical methods. Second, data on many of these processes is prohibitively expensive (or impossible) to obtain and so at most a handful of observations will exist in a given time – as in the case of a country's GDP or a company's sales. Hence, there’s a strong need to develop methods that can cope with these two difficulties.

The M4-Competition --the continuation of three previous ones organized by Spyros Makridakis-- is the latest challenge on time series forecasting for different types of time series \citep{MCompetition}. These competitions have attracted great interest in both the academic literature and among practitioners and have provided objective evidence of the most appropriate way of forecasting various variables of interest. The purpose of the M4-Competition is to replicate the results of the previous three ones and extend them in two directions. First, the number of series is increased to 100,000, and second, Machine Learning (Neural Network) forecasting methods are included.
 
The recurring result of the M Competitions is that simple methods outperform complex ones. This year's M4 competition generally reproduced this result with one exception: the winner of the competition, by a significant margin, was Slawek Smyl’s hybrid Exponential Smoothing-Recurrent Neural Network (ES-RNN) method – a synthesis of exponential smoothing techniques with recurrent nueral networks (RNN). \citep{Smyl2018M4Model}

Our task is to make state-of-the-art forecasting fast, accessible, and generalizable. First, we achieve fast training by porting Smyl's original C++ submission to Pytorch. This vectorization enables the use of a GPU, providing up to a 322x training speedup. Second, Python code makes the model more accessible to the forecasting and machine learning community.\footnote{Python is not only more popular (38.8\% vs 25.4\%) but also the \textit{most wanted language} for two years in a row \citep{StackOverflow2018Stack2018}.} Finally, the use of Pytorch allows easier generalization of the model as the library contains several architectures that could complement the ES-RNN for application in non-M4 time-series. We expect our contribution to speed up the adoption of hybrid models in time-series forecasting.

\section{Related Work}
\label{sec:relatedwork}

The field of time series analysis has not strongly deviated from statistical foundations laid in the 1950s. Seminal among these are the ARIMA method \citep{Box2008TimeControl} and the Exponential Smoothing method \citep{BrownExponentialDemand} \citep{Winters1960ForecastingAverages}. In addition to innovations resulting from the M4 competition, there has been research into using deep neural network (DNN) architectures for time series forecasting. The primary focus of this research has been to transfer techniques from other non-temporal DNN structures to time series forecasting.  

\cite{Qin2017APrediction} have added attention to the prediction algorithm by developing a dual-stage architecture. Others have incorporated learnings from the field of computer vision by implementing residual LSTMs \citep{KimResidualRecognition}. as well as dilated LSTMs \citep{ChangDilatedNetworks}. The latter having seen significant success in its CNN formulation \citep{YuMULTI-SCALECONVOLUTIONS}.

The popularity of DNNs in time series forecasting has also been verified by the development pre-trained weights that leverage a large number of time series. \citep{Malhotra2017TimeNet:Classification} This latter development has not only shown strong generalization but also has enabled DNNs to be applied to environments with relatively fewer training samples. The applications of such improvements in DNN, time series architectures has been leveraged heavily in a myriad of industries from predicting power demand \citep{Torres2019DeepPower} to predicting extreme events at Uber \citep{LaptevTime-seriesUberb}.

\section{Model Description}
\label{sec:modeldesc}

Our main contribution is a new ES-RNN implementation. In order to achieve our objective of a fast, accessible, and general forecasting engine, we re-engineer the Smyl's M4 Competition C++ submission for GPU computation. In contrast to the first version of ES-RNN, we exploit PyTorch's eager execution by initializing per-series parameters which are dynamically included in the computational graph.\footnote{We can see the parameters required for each series in Equations \ref{eq:level}, \ref{eq:trend} and \ref{eq:season}.} This technique enables us to vectorize ES-RNN and train on a GPU using the Nvidia CUDA library. The results in Section \ref{sec:results} show an important reduction in computation times. Furthermore, accessibility and generality are derived from the use of Python and PyTorch, respectively.

While the hybrid ES-RNN algorithm is discussed in detail in a blog post by \cite{Smyl2018M4Model}, we provide a condensed description of the learning process. The algorithm is divided into two distinct layers: a pre-processing layer that uses exponential smoothing and a LSTM layer that updates the per-series parameters of the Holts-Winter model.

\subsection{Pre-processing layer}
\label{subsec:prelayer}
In the pre-processing layer, Holts-Winter exponential smoothing \citep{Winters1960ForecastingAverages} with multiplicative seasonality and trend is applied via

\begin{align}
    l_t &= \alpha(\frac{y_t}{s_{t-m}}) + (1 - \alpha)l_{t-1} b_{t-1} \label{eq:level}\\
    b_t &= \beta(\frac{l_t}{l_{t-1}}) + (1- \beta) b_{t-1} \label{eq:trend} \\ 
    s_{t} &= \gamma \frac{y_t}{l_{t-1} b_{t-1}} + (1-\gamma)s_{t-m} \label{eq:season}\\
    \Hat{y}_{t+h} &= l_t  b_t^h s_{t-m+h^+_m} \label{eq:predEq}
\end{align}

where $l$ is a state variable for level, $b$ is a state variable for trend and $s$ is a multiplicative seasonality coefficient. $\alpha, \beta$ and $\gamma$ are smoothing coefficients between zero and one. An $h$ step forecast is given by Equation \ref{eq:predEq}.

\cite{Smyl2018M4Model} used constant size input and output windows. The output windows were defined by the prediction horizon. Conversely, the input window was determined heuristically as described in  \cite{Smyl2018M4Model}.

\cite{Smyl2018M4Model} further details how the Holt-Winters and the RNN methods were merged to create the final model. Since the model no longer considers local linear trend, Equation \ref{eq:trend} is replaced by an RNN from the model as follows:
\begin{align}
    \Hat{y}_{t+1 \dots t+h} &= RNN(X_t) * l_t * s_{t+1 \dots t+h} \label{eq:modifiedEq}\\
    x_i &= \frac{y_i}{l_t s_i} \label{eq:x}
\end{align}
$X_t$ is a vector of normalized, de-seasonalized features of which a scalar component $x_t$ is calculated via Equation \ref{eq:x}.

\subsection{Deep Learning Layer}
\label{subsec:nnlayer}

\begin{figure}[t]
\centering
  \includegraphics[width=\linewidth]{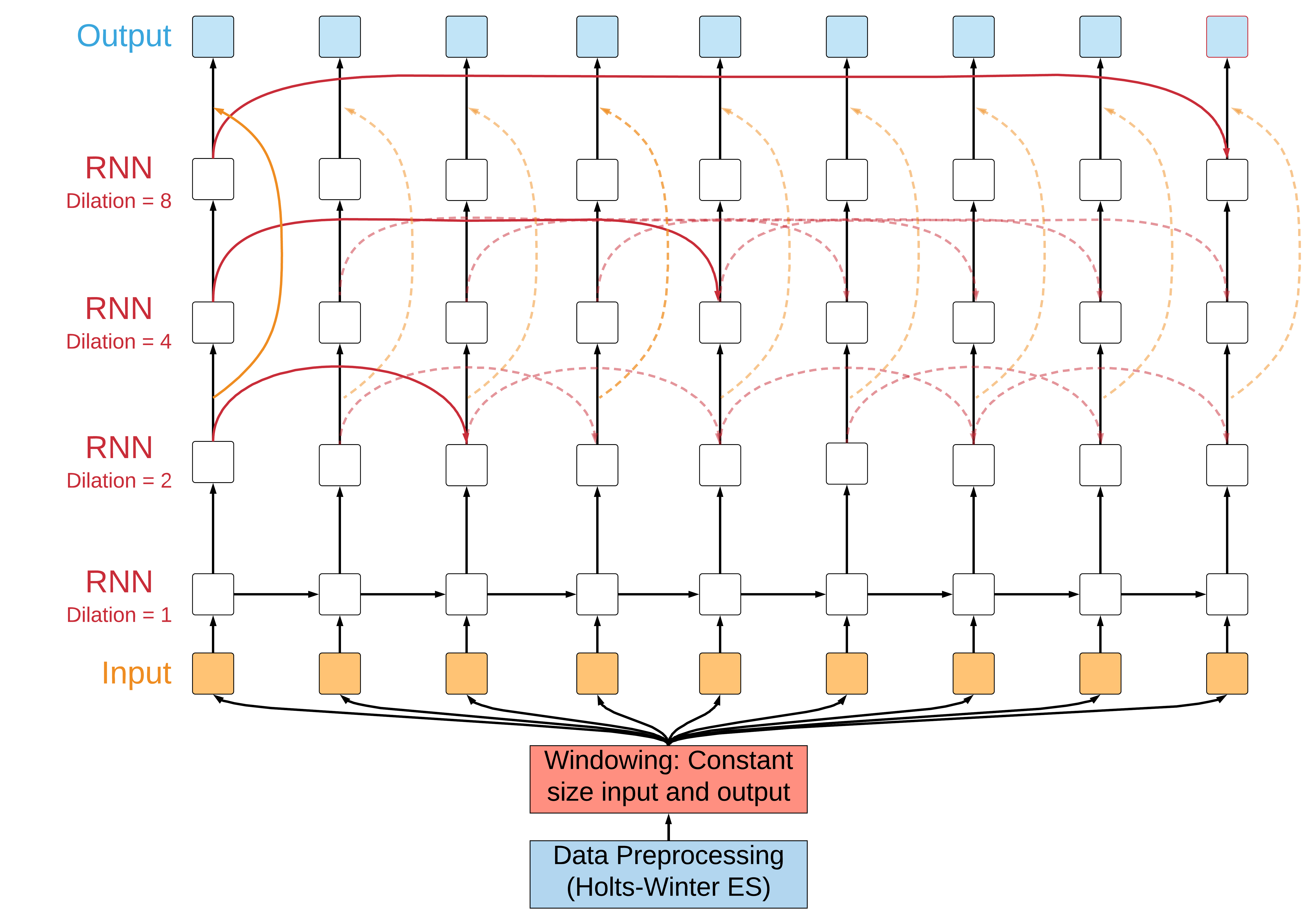}
  \caption{NN Architecture by \cite{Smyl2018M4Model}}
  \label{fig:slawekModels}
\end{figure}

The neural network architectures used by \cite{Smyl2018M4Model} are detailed in Figure \ref{fig:slawekModels}. A summary of the different models for quarterly, monthly and yearly time periods is presented in Table \ref{tab:nnparams}. The model at its base consists of a LSTM layer with skip connections to form the Dilated LSTM network as described in \cite{ChangDilatedNetworks}. The main advantages of this LSTM structure, when compared to a vanilla LSTM, are that it greatly increases computational efficiency and allows the network to \textit{remember} information from earlier time instances. For example, in Figure \ref{fig:slawekModels}, the RNN has dilations (1, 2) and (4, 8). This means that the first LSTM hidden weights are inputs to the next cell in the layer, however the second layer has a dilation of two which means that the hidden weights and the bias weights are forwarded two cells forward and so on and so forth. The RNN layers also have a residual connections that helps stabilize the network's training. Finally, we have a simple linear layer at the end for adapting the RNN output to the output prediction window in the form of \textit{normalized} and \textit{de-seasonalized} data, an important distinction which we discuss further in subsection \ref{subsec:testingnn}.

It is important to note that the RNN and the classical Holts-Winters parameters detailed in Section \ref{subsec:prelayer} are jointly trained. The power of \cite{Smyl2018M4Model} lies in the co-training of both the per-time series Holts-Winters parameters and the general RNN parameters.

\begin{table}[h]
\centering
\begin{tabular}{@{}lcc@{}}
\toprule
\multicolumn{1}{l}{\textbf{Time Frame}} & \textbf{Dilations} & \textbf{LSTM Size} \\ \midrule
Monthly                                 & (1, 3), (6, 12)    & 50                 \\
Quarterly                               & (1, 2), (4, 8)     & 40                 \\ 
Yearly                                   & (1, 2), (2, 6)    & 30                 \\ \bottomrule
\end{tabular}
\caption{Summary of network parameters}
\label{tab:nnparams}
\end{table}

\subsection{Training of the network}
\label{subsec:nntraining}

The training of this hybrid model is different from most neural network training. First, in order to have an initial estimate of the \textit{level} and \textit{seasonality} coefficients, we compute a primer estimate following the classical Holts-Winters equations  \ref{eq:level} and \ref{eq:season}. Note that if we have $N$ series in our dataset, the model will store $N * (2 + S)$ Holt-Winters parameters, where S is the length of the seasonality, as each series have its own \textit{level} and \textit{seasonality} smoothing parameters along with initial seasonality values ($S$).

In \cite{Smyl2018M4Model}'s CPU implementation, the network is trained per series However, the use of dynamic GPU computational frameworks such as PyTorch and, recently, Tensorflow enable batching. In Figure \ref{fig:slawekModels}, the input to the left most input box can be thought of as a matrix of \textit{Batch Size x Longest Input Window Length in Batch}. We will discuss further in Sections \ref{sec:datadesc} and \ref{sec:dataprep} why we need to pad and chunk each time-series into windows.

\subsection{Testing of the network}
\label{subsec:testingnn}

For generating the output of the network, we feed the output of the LSTM through a non-linear layer with a TanH activation function and finally through a final linear layer to get our predictions. This output is a \textit{de-seasonlized} and \textit{normalized} output which is not the case of our truth data. To re-seasonalized and de-normalize our outputs, we use the Holts-Winters equations and the parameter estimates we have for each series to arrive at the desired output. 

\subsection{Loss metrics}
\label{subsec:loss}

To calculate our loss, we first unpad and mask the output as we do not include these loss values in our calculation. The metrics used in the M4 competition were Symmetric Mean Absolute Percent Error (sMAPE) and Mean Absolute Scaled Error (MASE). Detailed explanations of these metrics are included in referenced post by \citep{Smyl2018M4Model}. However, since the metrics are non-differentiable, we employ a surrogate loss function called Pin-ball loss as defined in \citep{Takeuchi2006NonparametricEstimation}.

\section{Data Description}
\label{sec:datadesc}

The M4 competition data comprises 100K time-series with Yearly, Quarterly, Monthly, Weekly, Daily and Hourly data. Moreover, the dataset was sampled from real data -- categories encompass Demographic, Financial, Industrial, Macro- and Micro-economic and Other. The series counts broken out by frequency and type are in Table \ref{tab:countbytime}. However, the competition is more heavily focused on Yearly, Quarterly and Monthly data frequencies. We are of the opinion that this data is commonly generated from expensive or irreplicable processes, such as census or financial data.

\begin{table}[H]
\centering
\begin{tabular}{@{}cccccccc@{}}
\toprule
\multicolumn{1}{c}{\textbf{Frequency}} & \textbf{Demographic} & \textbf{Finance} & \textbf{Industry} & \textbf{Macro} & \textbf{Micro} & \textbf{Other} & \textbf{Total} \\ \midrule
Yearly                                 & 1,088                 & 6,519             & 3,716              & 3,903           & 6,538           & 1,236           & 23,000          \\
Quarterly                              & 1,858                 & 5,305             & 4,637              & 5,315           & 6,020           & 865            & 24,000          \\
Monthly                                & 5,728                 & 10,987            & 10,017            & 10,016          & 10,975          & 277            & 48,000          \\
Weekly                                 & 24                   & 164              & 6                 & 41             & 112            & 12             & 359            \\
Daily                                  & 10                   & 1,559             & 422               & 127            & 1,476           & 633            & 4,227           \\
Hourly                                 & 0                    & 0                & 0                 & 0              & 0              & 414            & 414            \\
Total                                  & 8,708                 & 24,534            & 18,798             & 19,402          & 25,121          & 3,437           & 100,000         \\ \bottomrule
\end{tabular}
\caption{M4 data by type and series frequency}
\label{tab:countbytime}
\end{table}

The dataset has two significant characteristics: First, the data is purely uni-variate with no guaranteed links between series other than broad sampling category. This lack of temporal or other connections between observations makes the forecasting challenge more difficult for complex methods as the sample itself is the only representation of signal. Second, each series is not only variable length but at times very small for an appropriate input window into the output forecasting horizon.

\section{Data Preparation}
\label{sec:dataprep}

\subsection{Validation Data Sets}
In time series analysis the prediction is defined as the most likely output given the series data. This means that to maximize the training data utilization we must use the latter part of our data for hold-out validation. We extracted the last $Output Size$ time steps from the end of our training set to create a validation set. Thus our data set takes the following form:

\begin{align}
    Train_{1...N - O*2-1}, Val_{N-O*2...N-O-1}, Test_{N-O...N}
    \label{cv_datasetup}
\end{align}

Where $O$ is the size of the forecast horizon and $N$ is the length of the dataset.

\subsection{Series Length Equalization} \label{subsec:seriesequ}
The series in the M4 data are variable length. To simplify the vectorization implementation, we equalize all series to a fixed length by frequency. We  further disregard all series below that specific length. In Section \ref{sec:futurework}, we discuss future work that will enable use of variable length series.

Key in determining an appropriate length to equalize series is the trade-off between removing too many series as a result of setting the minimum length too high or removing too much series history by making the minimum length too low. Hence, we visualized a histogram of all frequency series' length to understand what may be be best threshold by frequency. Table \ref{tab:freqseriesstats} shows statistics around series length in the M4 dataset. 

The heuristic we used to determine an appropriate threshold was one that maximized the data retention. This value usually fell somewhere in the second quartile. We used 72 as minimum series value for both quarterly and monthly time series frequencies. The cut-off value also provided several years of observed seasonality for both frequencies. 

\begin{table}[H]
\centering
\begin{tabular}{lrrrrrrr}
\toprule
\multicolumn{1}{c}\textbf{{}} &  \textbf{Mean} &    \textbf{Std-Dev} &    \textbf{Min} &    \textbf{25\%} &     \textbf{50\%} &     \textbf{75\%} &     \textbf{Max} \\ 
\midrule
Yearly    &    25 &       24 &    7 &   14 &    23 &    34 &   829 \\
Quarterly &    84 &       51 &    8 &   54 &    80 &   107 &   858 \\
Monthly   &   198 &      137 &   24 &   64 &   184 &   288 &  2776 \\
Weekly    &  1009 &      707 &   67 &  366 &   921 &  1590 &  2584 \\
Daily     &  2343 &     1756 &   79 &  309 &  2926 &  4183 &  9905 \\
Hourly    &   805 &      127 &  652 &  652 &   912 &   912 &   912 \\
\bottomrule
\end{tabular}
\caption{Frequency series length statistics}
\label{tab:freqseriesstats}
\end{table}

After equalizing the series length within frequency our full set of data is as follows:

\begin{align}
    Train_{N - O*2 - C...N - O*2-1}, Val_{N-O*2...N-O-1}, Test_{N-O...N}
    \label{cv_datasetup_postchop}
\end{align}
Where $C$ is the minimum length threshold.

\subsection{Data Windowing}
\label{subsec:data_wind}

\begin{figure}[h!]
\centering
  \includegraphics[width=\linewidth]{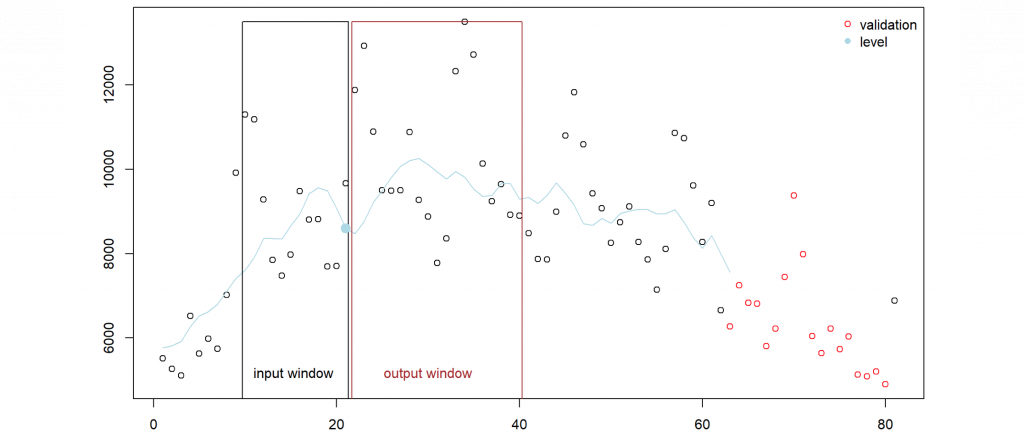}
  \caption{Time series normalization and deseasonalization as defined in \cite{Smyl2018M4Model}}
  \label{fig:slawek_time_series}
\end{figure}

Prior to passing the data through the second, RNN layer defined in section \ref{subsec:nnlayer}. We perform a windowing over the dataset, normalizing and deseasonalizing using the output from the ES layer defined in section \ref{subsec:prelayer}, as shown in Figure \ref{fig:slawek_time_series}.

In addition to the input window, we concatenate a one-hot representation of the time series category.

\section{Results}
\label{sec:results}
Ever since the first M competition, simple models such as Exponential Smoothing (ES) and ARIMA have outperformed more complex ones -- including machine learning. These type of models were also the benchmark for the M4 competition: a simple average of Simple, Holt and Damped ES models \cite{Makridakis2018TheForward} named Comb. The Comb model is a tough-to-beat benchmark, with a Rank of 19 in the M4 competition. Moreover, the second best submission by
\cite{Hyndman2018M4:Metalearning} in the M4 competition is a meta-learner that ensembles classical statistical model selection based on the R forecast package routines for automatic ARIMA, automatic ES, Theta, Naive, Seasonal Naive, Random Walk, TBATS from  \cite{DeLivera2010ForecastingSmoothing} and a shallow MLP. 

Table \ref{table:m4results} shows the results of the ES-RNN model by \cite{Smyl2018M4Model}, that of \cite{Hyndman2018M4:Metalearning} and our implementation of \cite{Smyl2018M4Model}'s ES-RNN on the GPU. Furthermore, Table \ref{tab:runtimes} shows the running times of our implementation on the GPU and the running time of \cite{Smyl2018M4Model} on CPU. Note that for monthly data, \cite{Smyl2018M4Model} were running the algorithm of 6 pairs of 2 workers and for quarterly data, 4 pairs of 2 workers were used.\footnote{We were not able to get the running time on yearly data from \cite{Smyl2018M4Model} at this time and we will update the table when we have their reports.} Finally, Table \ref{tab:smapebreakdown} breaks down the sMAPE achieved by our implementation in each of the frequencies (and corresponding models) for the six different data categories.

\begin{table}[h]
\centering
\begin{tabular}{@{}llllll@{}}
\toprule
\multicolumn{1}{c}{\multirow{2}{*}{\textbf{Model}}} & \multicolumn{4}{c}{\textbf{sMAPE by Frequency}} & \multicolumn{1}{c}{\multirow{2}{*}{\textbf{\% improvement}}} \\
\multicolumn{1}{c}{} & \multicolumn{1}{c}{\textbf{Yearly}} & \multicolumn{1}{c}{\textbf{Quarterly}} & \multicolumn{1}{c}{\textbf{Monthly}} & \multicolumn{1}{c}{\textbf{Average}} & \multicolumn{1}{c}{} \\ \midrule
Benchmark & 14.848 & 10.175 & 13.434 & 12.95 &  \\
\cite{Smyl2018M4Model} & \textbf{13.176} & \textbf{9.679} & 12.126 & 11.76 & 9.2\% \\
\cite{Hyndman2018M4:Metalearning} & 13.528 & 9.733 & 12.639 & 11.86 & 8.4\% \\
Our implementation & 14.42 & 10.09 & \textbf{10.81} & \textbf{11.50} & \textbf{11.2\%} \\ \bottomrule
\end{tabular}%
\caption{Comparison of results to the M4 baseline model}
\label{table:m4results}
\end{table}

\begin{table}[h]
\centering
\begin{tabular}{@{}lccl@{}}
\toprule
\multicolumn{1}{c}{\multirow{2}{*}{\textbf{Time Period}}} & \multicolumn{2}{c}{\textbf{\begin{tabular}[c]{@{}c@{}}Time taken for 15 epochs (mins)\end{tabular}}} & \multicolumn{1}{c}{\multirow{2}{*}{\textbf{Speed Improvement}}} \\
\multicolumn{1}{c}{} & \textbf{\cite{Smyl2018M4Model}} & \textbf{Our implementation} & \multicolumn{1}{c}{} \\ \midrule
Yearly & - & 9.33 & - \\
Quarterly & 2880 & 8.94 & \textbf{322x} \\
Monthly & 3600 & 31.91 & \textbf{113x} \\ \bottomrule
\end{tabular}%
\caption{Comparison of run-times for 15 epochs}
\label{tab:runtimes}
\end{table}

\begin{table}[h]
\centering
\begin{tabular}{@{}lccc@{}}
\toprule
\textbf{Data Category} & \textbf{Yearly} & \textbf{Quarterly} & \textbf{Monthly} \\ \midrule
Demographic & 11.6 & 10.78 & 6.31 \\
Finance & 15.86 & 10.74 & 11.58 \\
Industry & 19.57 & 7.44 & 12.38 \\
Macro & 15.68 & 9.57 & 12.45 \\
Micro & 11.35 & 11.63 & 9.94 \\
Other & 14.33 & 7.87 & 12.51 \\
Overall & 14.42 & 10.1 & 10.81 \\ \bottomrule
\end{tabular}
\caption{Breakdown of sMAPE by time period and category}
\label{tab:smapebreakdown}
\end{table}

\section{Discussion} 
\label{sec:discussion}

Most significant to this project is the increase in speed. We observe that our GPU implementation performs up to 322x faster on the quarterly dataset over the CPU approach. This speed up is driven by the vectorized implementation, which enables batching and parallelization.\footnote{Our implementation is in Pytorch, but is possible in any other deep learning framework with similar dynamic graph computation features.} Our implementation no longer requires running each time series data individually. As a result we could use batch sizes up to 2048 time series.

As shown in Table \ref{table:m4results}, \cite{Smyl2018M4Model} achieved superior results over the yearly and quarterly time interval. However, for monthly data our implementation surpasses the original model and benchmark by a strong margin. We believe there are two possible reasons for such a result. First, the longer training times of the original implementation likely made iteration over the parameter space prohibitive. Second, the series length equalization discussed in Subsection \ref{subsec:seriesequ} likely simplified the problem  to the model in some instances and omitted harder, shorter sequences in others. Overall, our implementation has the lowest weighted sMAPE score due to the mentioned improvement on the monthly dataset. 

We are pleased to see that our implementation is close to the M4 implementation for most frequencies. However, the large disparity in scores achieved by \cite{Smyl2018M4Model} for the yearly dataset could be attributed to the original author's use of attentive LSTM on top of the residual and dilated LSTM, which our current implementation does not include. Furthermore, \cite{Smyl2018M4Model} did not use any seasonality parameters for the yearly data which our implementation was not designed to handle. 

\section{Future Work/Extensions}
\label{sec:futurework}

\subsection{Variable Length Series}
\label{subsec:variableseries}

In section \ref{subsec:seriesequ}, we discussed how we equalized series length by frequency to simplify batch implementation. The next step is to support variable length series. This feature is critical as results both improve with the addition of more data and weaken as shorter series are introduced. The result of these competing effects will better inform whether batching is appropriate for this model structure

We anticipate this feature will require the use of padded sequences together with masking just before the loss function aggregation. This mask will zero those gradients that are derived from padded observations outside the series' real data. We recognize that there is risk to this approach as series reaches the frequency maximum length as padding will dramatically increase the number of required iterations when performing the exponential smoothing pre-processing. Therefore further development in this area may require a smaller batch size to hedge risk against unnecessary computation. 

\subsection{Multiple Seasonality} 
\label{subsec:multseas}

Smyl's M4 submission uses more than one seasonality. This is particularly important in the hourly frequency as the data may follow both a 24 and 168- hour cycles. \cite{Gould2008ForecastingPatterns} demonstrated that multiplicative multiple seasonality is achieved by maintaining a second set of seasonality smoothing and initial values, and when de-seasonalizing the data dividing by the corresponding seasonalities one-after-the-other. 

Secondarily, the model needs to be tuned to handle no seasonality in situations where no obvious cycle exists (i.e. yearly data).

\subsection{Hyper-parameter Tuning}
One major purpose of this work was to validate that the primary components of ES-RNN can be vectorized and ported to dynamic frameworks. As a result, we kept all of Smyl's M4-winning parameters wherever possible. However, we don't know which parameters were set out of necessity given the CPU training structure and which were actually tuned in preparation for the competition. For example, in the winning submission, the hidden size for all RNNs was no more than 50. As a result, our GPU utilization was very low relative to other deep learning tasks that are often tuned to values in the hundreds. 


\subsection{Additional Penalization}
The M4-winning, DyNet/C++ model includes several layers of loss penalization that were not included in our work. 

First, as the RNN is modeling the series trend (i.e. the change in level from $t$ to $t+1$), the likelihood of  over-fitting could be reduced by penalizing dramatic changes in the level. This change would causes the model to favor a smoother forecast over a more variant forecast ceteris paribus.

Second, \cite{KruegerREGULARIZINGACTIVATIONS} have shown that penalizing abrupt movements in the hidden state can encourage a model to be more stable over long forecast horizons. Rather than penalizing the change from $t$ to $t+1$, Smyl's model penalized the average squared magnitude of the first layer's cell state within each stack. These penalties applied to the internals of the model favor a model that will likely be more stable after appropriate training, and could improve ES-RNN performance.

\subsection{Additional Frequencies}
\begin{figure}[t]
\centering
  \includegraphics[width=0.8\linewidth]{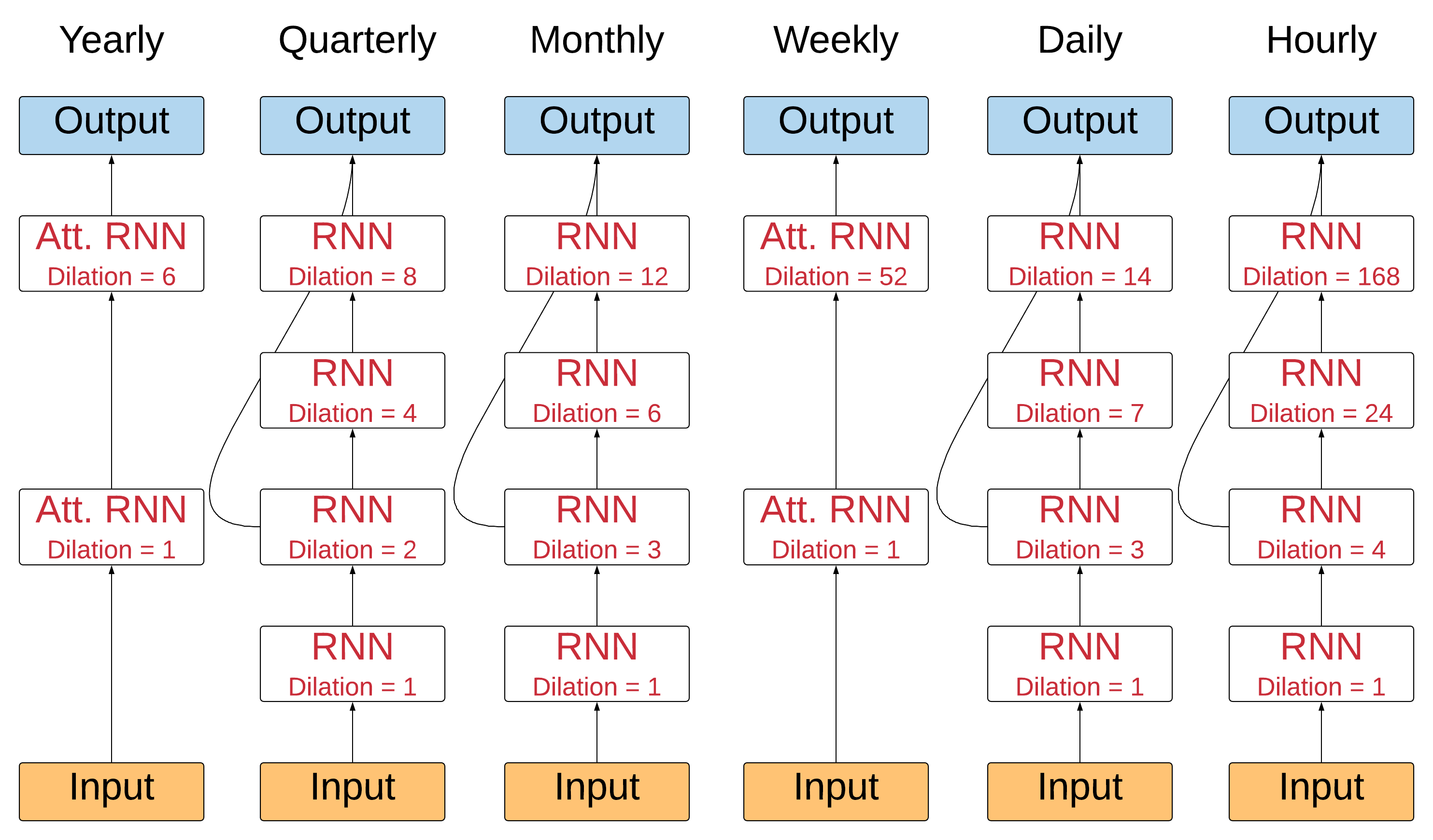}
  \caption{NN Architecture by \cite{Smyl2018M4Model}}
  \label{fig:modelstructs}
\end{figure}

We built out vectorized, ES-RNN functionality for a subset of M4 competition dataset as a proof of concept, but our approach will hypothetically scale to other time series provided we overcome the challenges presented by section \ref{subsec:multseas} as well as implement the other network structures Smyl used for the various frequencies shown in Figure \ref{fig:modelstructs}

Noteworthy differences between already the implemented quarterly and monthly frequencies and yearly, weekly, daily and hourly frequencies are the use of attention in yearly and weekly and the multiple seasonalities defined in hourly series. (Note: Daily is the same structure as quarterly and monthly).

\section{Conclusion}

In this project, we successfully implement the state-of-the-art ES-RNN algorithm in a fast, accessible, and generalizable forecasting framework. The main challenge we overcame was the training of per-time series parameters. This proved difficult due to the direct implementation of the original submission on a CPU. Our work focused on the vectorization of the per-time series parameters to enable GPU computation in a framework that supports eager execution (e.g. Pytorch). We obtained similar results to those of the original submission, but with orders of magnitude less training time. We anticipate that our contribution will enable a strong adoption of state-of-the-art algorithms over univariate series, and facilitate the generalization of the model towards specific problems where covariates are available.

\bibliography{references}

\begin{thebibliography}{19}
\providecommand{\natexlab}[1]{#1}
\providecommand{\url}[1]{\texttt{#1}}
\expandafter\ifx\csname urlstyle\endcsname\relax
  \providecommand{\doi}[1]{doi: #1}\else
  \providecommand{\doi}{doi: \begingroup \urlstyle{rm}\Url}\fi

\bibitem[MCo()]{MCompetition}
{M⁴ Competition – The M⁴ (Makridakis 4) Forecasting Competition}.
\newblock URL \url{https://www.m4.unic.ac.cy/}.

\bibitem[Box et~al.(2008)Box, Jenkins, and Reinsel]{Box2008TimeControl}
George E.~P. Box, Gwilym~M. Jenkins, and Gregory~C. Reinsel.
\newblock \emph{{Time series analysis : forecasting and control}}.
\newblock John Wiley, 2008.
\newblock ISBN 9781118619193.

\bibitem[Brown()]{BrownExponentialDemand}
Robert~B. Brown.
\newblock {Exponential Smoothing for Predicting Demand}.
\newblock URL
  \url{https://www.industrydocumentslibrary.ucsf.edu/tobacco/docs/#id=jzlc0130}.

\bibitem[Chang et~al.()Chang, Zhang, Han, Yu, Guo, Tan, Cui, Witbrock,
  Hasegawa-Johnson, and Huang]{ChangDilatedNetworks}
Shiyu Chang, Yang Zhang, Wei Han, Mo~Yu, Xiaoxiao Guo, Wei Tan, Xiaodong Cui,
  Michael Witbrock, Mark Hasegawa-Johnson, and Thomas~S Huang.
\newblock {Dilated Recurrent Neural Networks}.
\newblock Technical report.
\newblock URL \url{https://github.com/code-terminator/DilatedRNN}.

\bibitem[De~Livera et~al.(2010)De~Livera, Hyndman, and
  Snyder]{DeLivera2010ForecastingSmoothing}
Alysha~M De~Livera, Rob~J Hyndman, and Ralph~D Snyder.
\newblock {Forecasting time series with complex seasonal patterns using
  exponential smoothing}.
\newblock Technical report, 2010.
\newblock URL \url{http://www.buseco.monash.edu.au/depts/ebs/pubs/wpapers/}.

\bibitem[Gould et~al.(2008)Gould, Koehler, Ord, Snyder, Hyndman, and
  Vahid-Araghi]{Gould2008ForecastingPatterns}
Phillip~G. Gould, Anne~B. Koehler, J.~Keith Ord, Ralph~D. Snyder, Rob~J.
  Hyndman, and Farshid Vahid-Araghi.
\newblock {Forecasting time series with multiple seasonal patterns}.
\newblock \emph{European Journal of Operational Research}, 191\penalty0
  (1):\penalty0 207--222, 11 2008.
\newblock ISSN 0377-2217.
\newblock \doi{10.1016/J.EJOR.2007.08.024}.
\newblock URL
  \url{https://www.sciencedirect.com/science/article/pii/S0377221707008740?via%3Dihub}.

\bibitem[Hyndman(2018)]{Hyndman2018M4:Metalearning}
Rob Hyndman.
\newblock {M4: Forecasting Metalearning}, 2018.
\newblock URL
  \url{https://github.com/robjhyndman/M4metalearning/blob/master/docs/metalearning_example.md}.

\bibitem[Kim et~al.()Kim, El-Khamy, and Lee]{KimResidualRecognition}
Jaeyoung Kim, Mostafa El-Khamy, and Jungwon Lee.
\newblock {Residual LSTM: Design of a Deep Recurrent Architecture for Distant
  Speech Recognition}.
\newblock Technical report.
\newblock URL \url{https://arxiv.org/pdf/1701.03360.pdf}.

\bibitem[Krueger and Memisevic()]{KruegerREGULARIZINGACTIVATIONS}
David Krueger and Roland Memisevic.
\newblock {REGULARIZING RNNS BY STABILIZING ACTIVATIONS}.
\newblock Technical report.
\newblock URL \url{https://arxiv.org/pdf/1511.08400.pdf}.

\bibitem[Laptev et~al.()Laptev, Yosinski, Erran~Li, and
  Smyl]{LaptevTime-seriesUberb}
Nikolay Laptev, Jason Yosinski, Li~Erran~Li, and Slawek Smyl.
\newblock {Time-series Extreme Event Forecasting with Neural Networks at Uber}.
\newblock Technical report.
\newblock URL
  \url{http://roseyu.com/time-series-workshop/submissions/TSW2017_paper_3.pdf}.

\bibitem[Makridakis et~al.(2018)Makridakis, Spiliotis, and
  Assimakopoulos]{Makridakis2018TheForward}
Spyros Makridakis, Evangelos Spiliotis, and Vassilios Assimakopoulos.
\newblock {The M4 Competition: Results, findings, conclusion and way forward}.
\newblock \emph{International Journal of Forecasting}, 34\penalty0
  (4):\penalty0 802--808, 10 2018.
\newblock ISSN 0169-2070.
\newblock \doi{10.1016/J.IJFORECAST.2018.06.001}.
\newblock URL
  \url{https://www.sciencedirect.com/science/article/pii/S0169207018300785}.

\bibitem[Malhotra et~al.(2017)Malhotra, TV, Vig, Agarwal, and
  Shroff]{Malhotra2017TimeNet:Classification}
Pankaj Malhotra, Vishnu TV, Lovekesh Vig, Puneet Agarwal, and Gautam Shroff.
\newblock {TimeNet: Pre-trained deep recurrent neural network for time series
  classification}.
\newblock 6 2017.
\newblock URL \url{http://arxiv.org/abs/1706.08838}.

\bibitem[Qin et~al.(2017)Qin, Song, Chen, Cheng, Jiang, and
  Cottrell]{Qin2017APrediction}
Yao Qin, Dongjin Song, Haifeng Chen, Wei Cheng, Guofei Jiang, and Garrison
  Cottrell.
\newblock {A Dual-Stage Attention-Based Recurrent Neural Network for Time
  Series Prediction}.
\newblock 4 2017.
\newblock URL \url{http://arxiv.org/abs/1704.02971}.

\bibitem[Smyl et~al.(2018)Smyl, Ranganathan, and Pasqua]{Smyl2018M4Model}
Slawek Smyl, Jai Ranganathan, and Andrea Pasqua.
\newblock {M4 Forecasting Competition: Introducing a New Hybrid ES-RNN Model},
  2018.
\newblock URL \url{https://eng.uber.com/m4-forecasting-competition/}.

\bibitem[{Stack Overflow}(2018)]{StackOverflow2018Stack2018}
{Stack Overflow}.
\newblock {Stack Overflow Developer Survey 2018}, 2018.
\newblock URL
  \url{https://insights.stackoverflow.com/survey/2018/#most-popular-technologies}.

\bibitem[Takeuchi et~al.(2006)Takeuchi, Le, Sears, and
  Smola]{Takeuchi2006NonparametricEstimation}
Ichiro Takeuchi, Quoc~V Le, Timothy~D Sears, and Alexander~J Smola.
\newblock {Nonparametric Quantile Estimation}.
\newblock Technical report, 2006.
\newblock URL \url{http://www-stat.stanford.edu/ElemStatlearn.}

\bibitem[Torres et~al.(2019)Torres, Troncoso, Koprinska, Wang, and
  Mart{\'{i}}nez-{\'{A}}lvarez]{Torres2019DeepPower}
J.~F. Torres, A.~Troncoso, I.~Koprinska, Z.~Wang, and
  F.~Mart{\'{i}}nez-{\'{A}}lvarez.
\newblock {Deep Learning for Big Data Time Series Forecasting Applied to Solar
  Power}.
\newblock pages 123--133. Springer, Cham, 6 2019.
\newblock \doi{10.1007/978-3-319-94120-2{\_}12}.
\newblock URL \url{http://link.springer.com/10.1007/978-3-319-94120-2_12}.

\bibitem[Winters and R.(1960)]{Winters1960ForecastingAverages}
Peter~R. Winters and Peter R.
\newblock {Forecasting Sales by Exponentially Weighted Moving Averages}.
\newblock \emph{Management Science}, 6\penalty0 (3):\penalty0 324--342, 4 1960.
\newblock ISSN 0025-1909.
\newblock \doi{10.1287/mnsc.6.3.324}.
\newblock URL \url{http://pubsonline.informs.org/doi/abs/10.1287/mnsc.6.3.324}.

\bibitem[Yu and Koltun()]{YuMULTI-SCALECONVOLUTIONS}
Fisher Yu and Vladlen Koltun.
\newblock {MULTI-SCALE CONTEXT AGGREGATION BY DILATED CONVOLUTIONS}.
\newblock Technical report.
\newblock URL \url{https://arxiv.org/pdf/1511.07122.pdf}.

\end{thebibliography}
\bibliographystyle{11785_project}

\end{document}